# CAN AN ORGANISM ADAPT ITSELF TO UNFORESEEN CIRCUMSTANCES?

Alexey V. Melkikh

*Ural State Technical University, Ekaterinburg, 620002, Mira str., 19, Russia,*

[mav@dpt.ustu.tu](mailto:mav@dpt.ustu.tu)*, 7(343)374-78-54*

**Abstract.** A model of an organism as an autonomous intelligent system has been proposed. This model was used to analyze learning of an organism in various environmental conditions. Processes of learning were divided into two types: strong and weak processes taking place in the absence and the presence of aprioristic information about an object respectively. Weak learning is synonymous to adaptation when aprioristic programs already available in a system (an organism) are started. It was shown that strong learning is impossible for both an organism and any autonomous intelligent system. It was shown also that the knowledge base of an organism cannot be updated. Therefore, all behavior programs of an organism are congenital. A model of a conditioned reflex as a series of consecutive measurements of environmental parameters has been advanced. Repeated measurements are necessary in this case to reduce the error during decision making.

## Introduction

The ability of animals to learn, that is, react adequately to the appearance of an unknown image in their sight, is considered as one of basic properties of living systems (see, for example, (McFarland, 1985)), which was modeled time and again (Bush & Mosteller, 1955, Staddon, 1983, Bergman & Feldman, 1995, Dukas, 1998, Kerr & Feldman, 2003). It is generally agreed that some part of information about the state of an organism is encoded in genes (is congenital), while the other part is acquired during learning by experience. However, although models of living systems are many, mechanisms of learning are far from being clear.

In particular, a detailed analysis of the means by which an organism receives information about the environment and makes decisions on adequate actions reveals a contradiction.

On the other hand, development of a self-learning system capable of acquiring knowledge from the environment is one of central problems in the theory of the artificial intelligence. Some researchers think that such systems already exist or will be created in the nearest future (cognitive computers, see, for example, (Brachman, 2001)). However, the analysis of the state of the art with learning systems shows that neither of them can set itself new tasks and adapt to an arbitrary environment (Andrew, 1983, Papert, 1980, Spier & McFarland, 1997, Gupta & Sinha, 1996, White & Sofge, 1992). Models of the adaptive behavior of agents should also a priori include all properties, which these agents may ever possess (for example, (Maturana & Varela, 1980, Rich & Knight, 1991, Iizuka & Ikegami, 2004)). New properties of agents do not appear in this case. Is this a temporary drawback of modern systems or such a system cannot be created in principle?

At the same time, frequently the notions "learning" and "adaptation" are not clearly defined, leading to misunderstanding.

The present paper deals with an alternative model of the behavior of organisms and intelligent systems, which is based on the operation of aprioristic programs.

### 1. What do the terms "learning" and "adaptation" imply?

The notion "learning" is used in a number of sciences: information science, psychology, physiology, ethology, etc. Hence, this notion is defined differently (see, for example, (Luger, 2003)). The ability of an intelligent system for learning is frequently taken as its immanent property whose nature is not discussed (Cohen et al., 1990, Curran & Keele, 1993).

The term "adaptation" (when a system chooses some program from the programs available at its disposal) is also used in the literature. It is often taken as a synonym of the notion "learning" (the adaptive behavior, an adaptive automat, (Hauert & Stenull, 2002, Delgado & Sole, 2000, Staddon, 1983, Frank, 1997) etc.).

The adaptive behavior of agents is formalized beginning, mainly, from lower hierarchy levels. It was shown (Jonker et al., 2002) that all biochemical processes occurring in an elementary organism can be described in a high-level language.

An important property of the adaptive behavior is its purposefulness, which becomes apparent at different levels of the organization of the living matter. In this hierarchy, purposes of lower levels are directly connected with the interaction between an organism and the environment in the present situation, while purposes of higher levels determine its long-term behavior (Burtsev, 2004). What is the source of purposes then? Are they preset a priori or do they appear during vital activities of an organism? What is the mechanism of their appearance in the latter case?

The notion of "learning" has been formalized most in the theory of the artificial intelligence. However, "learning" is understood differently in different sections of the artificial intelligence.

For example, Herbert Simon defined "learning" as "any change in a system, which improves the solution of a task upon its repeated presentation or leads to the solution of another task on the basis of the same data" (Luger, 2003). However, from this definition it is not clear if it implies a system having an invariable internal structure or a system whose structure can be changed at will. If, for example, the man interferes directly and creates actually another system instead of an existing system, will this be learning? Such a process obviously has nothing to do with learning, but is related to another domain. Therefore, it is necessary to elaborate on the type of the interaction between an intelligent system and the environment.

From the analysis of the literature on the artificial intelligence it is possible to distinguish two types of learning depending on whether an intelligent system or an organism has or has not *aprioristic information* about an object. Aprioristic information implies information about an object, which a system had before presentation of the object.

Let us refer to learning of the first (strong) type as learning in the absence of aprioristic information about an object and learning of the second (weak) type as learning in the presence of this information.

All modern systems of the artificial intelligence are capable of learning only after the second type, i.e. in the presence of aprioristic information. Indeed,

1. Heuristic methods of problem solving are based on aprioristic information about the object domain comprising the problem. The heuristics may prove to be incapable of finding the solution altogether. This limitation cannot be removed by the best heuristics (Garey & Johnson, 1979). The key question is whence the heuristics comes? How to develop a new heuristics? The theory does not answer this question.

2. The recursive search represents a natural method for realization of such strategies of the artificial intelligence as the graph search. However, an exact objective should be set for this method to be used. If objects (even one object) are not defined, the recursive procedure cannot be applied (Luger, 2003).

3. Production systems are used for the conflict resolution. In this case, it is necessary to a priori have a standard sample

(pattern), which determines the possibility to use rules of production systems (Luger, 1994).

4. Expert systems rest on the same rules (Luger, 2003). The core of an expert system is a knowledge base, which contains knowledge from a particular applied domain. Knowledge in an object domain determines and updates an expert database (Minsky, 1987). Is it possible to imagine a system, which acts as an expert for itself? Obviously not, because a code can be corrected only when correct knowledge is known.

5. Machine learning (Luger, 2003), be it symbolic, neuronet or emergent (genetic algorithms) learning, is based on the presentation of a priori preset training examples. Indeed, learning of a neuron network implies the assignment of relation weights between neurons, which is realized in a special regime (but not in the pattern recognition regime). In this case, the person, who assigns standard templates to this neuron network, has aprioristic information about the object. One more problem is how to determine that the system has come to the correct decision.

Is learning of the first (strong) type possible? Can an organism (a system) update the knowledge base itself and adapt to unforeseen circumstances?

To answer this question, we shall consider the model of an organism as an intelligent system.

2. **The model of an organism as an autonomous intelligent system**

Let us define first the class of systems, which may be posed the problem of learning in the absence of aprioristic information about an object. The major property of intelligent systems, which allows this classification, is the method of their interaction with the environment.

By this method, all intelligent systems can be divided into two classes: systems acquiring all information about the environment through measurements (pattern recognition) and those exposed to random effects when unrecognized information can be imported to a system. In the first case, the system is logically closed and its internal rules cannot be changed by a random external effect. In the second case, the system is not logically closed and its rules and structure can be changed randomly. An example is a machine assembled in a production line. If the man or another intelligent system decides to change the machine design, the machine may be imparted any property for solution of any problem. Then the "machine + external system" entity will be logically closed.

This definition is similar to the notion "an autonomous system" (see, for example, (Ruiz-Mirazo & Moreno, 2004)). However, the said notion has not been defined clearly enough in the literature.

The presence of an uncontrollable external effect makes the optimization problem insoluble, because any degree of freedom of a system and, therefore, any internal operation and an object of the system can be changed arbitrarily. The effect of other intelligent systems may update the database of a system and lead to a random change of its behavior. However, this system will not be autonomous. In this case, the problem of learning is trivial: the process is fully controlled from the outside by another intelligent system. Therefore, the question about the source of new knowledge is just re-applied to another system. Then the combined system will be autonomous.

The analysis of the behavior of organisms demonstrated that they belong to autonomous systems in the aforementioned sense: their internal degrees of freedom (the nervous system and other control systems) are protected from direct outside effects. The state of internal degrees of freedom (neurons) changes thanks to measurements of environmental conditions, that is, the pattern recognition by receptors.

The present study deals just with the first type of systems, which acquire information from the outside by the pattern recognition.

So, let us consider an organism as a logically closed intelligent system (Fig. 1) having the following properties:

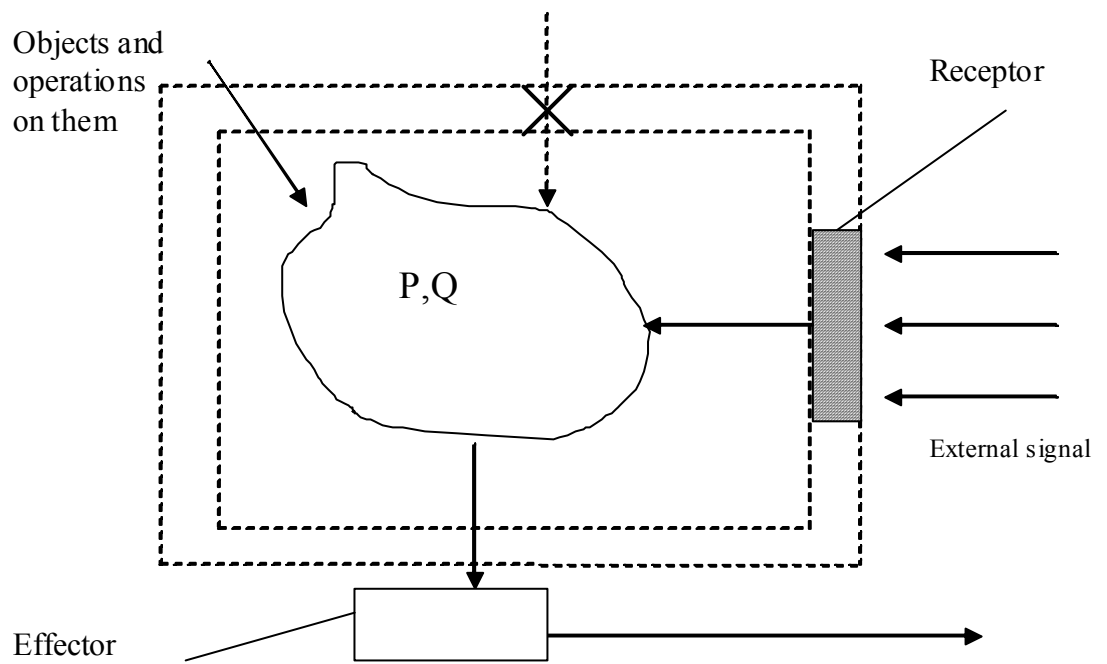

Fig.1. Model of an organism as an autonomous intelligent system

Property 1: A system includes internal independent objects (language words) $Q_1 \ldots Q_n$. The objects may be subject to operations $P_1 \ldots P_m$. Since the system is logically closed, neither objects nor operations on them can be changed due to actions from the outside. New objects cannot be created either by applying operations to a set of assigned objects, because in this case a new object will not be independent.

Property 2: A system has a receptor for acquisition of information about environmental conditions. A signal, which is received by the receptor, is compared with internal objects available in the system (patterns are recognized). This recognition can be pictured as a tree (a graph).

Property 3: Some operations on the objects start effectors, i.e. the system performs operations on the environment. The structure of the system is such that all operations lead to the solution of one of a priori defined tasks $\Sigma_l(Q_i, P_i)$. The quality of the solution of each task can be estimated by the quality functional $\Phi_l(Q_i, P_i)$, which is the larger, the better the task is solved. We shall assume that the structure of the system provides the condition of the maximum $\Phi_l(Q_i, P_i) \to \max$.

Property 4: A system has a memory, which may store results of previous measurements or intermediate operations.

The class of the intelligent systems at hand is very wide. It includes all living organisms (including the man), the artificial intelligence and adaptive systems.

## 3. Self-learning of a system in the absence of aprioristic information leads to a contradiction

Let us consider a situation when a system sees a pattern, which it can register (measure), but cannot recognize. Also, the system has not any aprioristic information about the pattern.

In this case, the recognition scheme can be pictured as follows (Fig. 2):

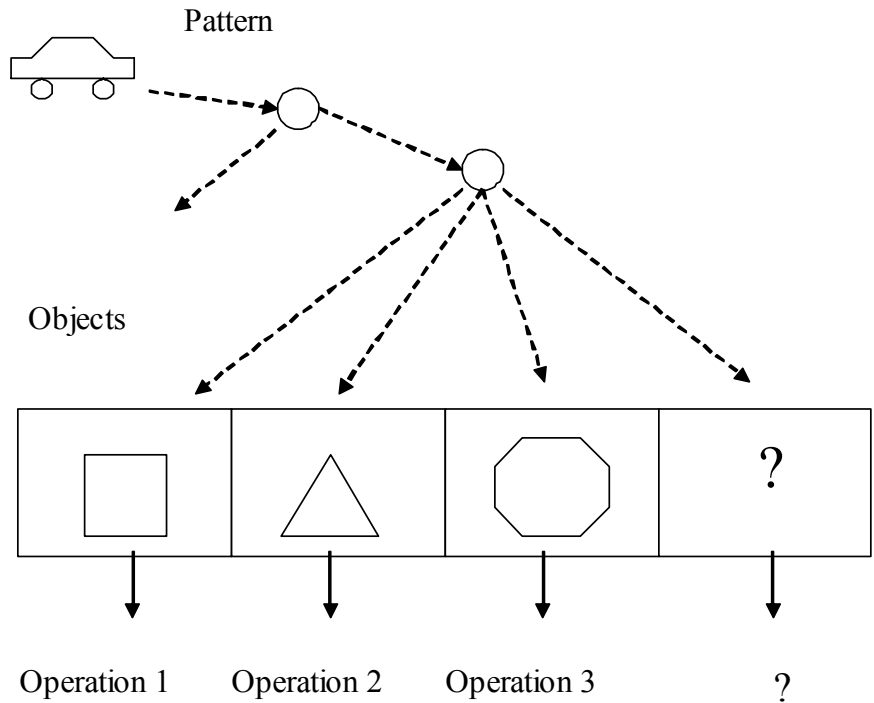

Fig.2. Tree of the recognition and decision making when an unknown object appears in the field of vision

Statement 1: If an intelligent system possessing properties 1-4 catches sight of a pattern Ω, which is not equivalent to any of the internal objects of the system, the effector cannot perform operations and actions, which would lead to the solution of any new task $\Sigma_{l+1}(\Omega, Q_i, P_i)$ related to this pattern.

The proof:

Two cases are possible:

1. The pattern Ω is not equivalent to any of $Q_i$ (the pattern does not belong to the set of internal objects of the system). Then neither operation on the pattern is defined. Effectors will not be started. Since all tasks, which the system can solve, are determined only by the internal objects and operations on them (property 3), the task list will not be updated.

2. Let the pattern be partially identified and prove to be equivalent to an object in the tree node $Q_i$. The further identification on the tree failed. In this case, the effector corresponding to the object $Q_i$ will be started. Then the object-defined operations can only be applied to the pattern. Therefore, in this case the task list will not be updated either. The statement has been proved.

The problem of updating the knowledge base of the system has a similar solution. In terms of the proposed model, the knowledge base represents a set of objects, actions on them, and outcomes of those actions (the quality functional).

<u>Statement 2</u>: An intelligent system, which possesses properties 1-4, cannot update its knowledge base.

<u>The proof</u>: A set of objects and operations on them cannot be updated by definition (property 1), because the objects are independent and operations on unrecognized patterns are not defined. The statement has been proved.

Thus, we have come to the conclusion that a system (an organism) cannot pose new tasks when an unrecognized object appears in the environment. The generation of new tasks or updating of the knowledge base by the system contradicts its properties.

### 3. The trial and error method

Let us dwell on a widespread mechanism of learning, namely learning by the trial and error method. This mechanism was discussed in the literature more than once (Beal, 2003, Minsky, 1987, Sloman, 1993).

Some authors use the term "casual learning" in this sense, assuming that possible actions are searched randomly.

It is easy to see however that this algorithm coincides with the heuristic method of the task solution (Luger, 2003) and, as it was shown in the foregoing, can be applied only when an intelligent system has aprioristic information about an object. (If this information is absent, nothing implies that the task will be solved by chance.) In this case, learning refers to the second (weak) type. Indeed, if a system (an organism) has an operation on an unrecognized pattern leading to the solution of at least one task, then (according to property 3) the unrecognized pattern should belong to the set of internal objects of the system. That is, the system includes a truncated mechanism of the recognition when the tree of objects contains a smaller number of edges and tops.

An important conclusion follows from the above reasoning: all behavior programs of organisms (including the man) are congenital. This conclusion contradicts today's theoretical concepts that only part of behavior programs of animals is laid in genes, while the rest of the programs are acquired during learning.

So, just one of the aprioristic programs, which suits best a given situation, is chosen during learning of the man and animals.

The above reasoning does not mean in the least that the environment plays an insignificant role in the behavior of animals and the man. Experiments with enzygotic twins demonstrated (Thompson et al., 2002) that the role of the environment is considerable. Of course, this statement does not contradict the main conclusion of the present study that all behavior programs of organisms are congenital. With respect to twins, for example, this means the following: although both twins have the same congenital programs of behavior, some programs will be started for one of the twins and other programs for the other (if the latter is in another environment).

## 4. Transfer of Experience from Parents to Descendants

Let us consider the case when an organism receives new information from its parents or other animals. Can it adapt itself to new conditions in this case? Considering what has been said above, the answer is negative.

It is important to note that in this case too the organism receives information via receptors. Therefore, any other organism (a member of the pack, the parent, etc.) is interpreted as part of the environment. Consequently, the aforementioned "*measurement - recognition – decision making*" scheme holds. Whichever actions a parent makes, they can be adequately interpreted only in one case: when the signal has been recognized, that is, has been compared with a standard sample. This comparison is possible in turn only if the standard sample has already been available in the organism (has been congenital).

What is the role of the experience transfer between animals? This role obviously reduces to triggering of aprioristic (congenital) programs of behavior. Indeed, if the environment has uncertainties, an important question is which programs to start. The start of behavior programs, which are not adequate to changes in the environment, can be catastrophic for an organism. Therefore, we shall consider a situation (which is most frequent) when errors occur during reception of the environment.

## 5. Repeated measurements as the basis of a conditioned reflex

Why then a certain type of the behavior is established not immediately after appearance of a new pattern, but only after its repeated occurrence? This can be easily explained if one considers that the quality functional of actions depends on the number of measurements. On the one hand, measurements require consumption and, on the other hand, the measurement error decreases as the number of measurements grows. Therefore, the quality functional of measurements can have an extremum (Melkikh, 2005). In other words, a system (an organism), which is in a complex environment, may find it unfavorable to immediately change its trajectory as changes occur in the environment, but prefers to perform a set of measurements and only then alter its behavior. More detailed calculations are given in a paper by Melkikh (2005).

Thus, repeated measurements of environmental conditions by a system are intended to reduce the error during operation of aprioristic programs. The problem of the acquisition of new information during formation of a conditioned reflex is covered in more detail elsewhere (Melkikh, 2002). It was shown that:

- an organism should recognize a pattern to form a conditioned reflex,
- an organism should have a program for the work with the pattern, which, in particular, should instruct that an effector should be started after a given pattern recurs a certain number of times,
- in this sense, conditioned reflexes do not differ fundamentally from unconditioned ones. The only difference is the number of pattern presentations,
- in this case, new programs cannot be created.

**6. The algorithm of the organism behavior under uncertainty conditions**

We shall use the aforementioned properties of an organism as an autonomous intelligent system and construct a model of the behavior of an organism in conditions of environmental uncertainty. In this case, the algorithm of actions of an organism can be presented as a set of procedures:

1. Measure (X,n)→Q. The measure procedure has a pattern of the environment at the input and the object, to which the external pattern was identified in measurements, at the output.
2. Do (P,Q). The action procedure (the control is transferred to effectors).
3. Memory (P,Q). The procedure of recording to the memory and memory read.
4. DoWhile($\Phi > \Phi_0$). The procedure of comparing the quality functional of the action with its maximum value. As a result, aprioristic programs are ordered.
5. Random(P,Q). The procedure of a random selection of an aprioristic program.

## Conclusion

It was shown that an organism cannot adapt itself to unforeseen circumstances, solve new tasks, update its knowledge base, or learn upon presentation of an unknown object. All behavior programs of the man and animals are congenital and cannot be acquired from learning. A conditioned reflex can be modeled as a series of consecutive measurements, which is intended to reduce the error during decision making. In this case, an organism does not develop new programs.

## References

Andrew, A.M. (1983). *Artificial intelligence*, Viable Systems Chillaton, Devon (U.K.). Abacus Press.

Beal, J. (2003). Leveraging learning and language via communication bootstrapping. *AI Memo*. 2003-007.

Bergman, A. & Feldman, M.W. (1995). On the evolution of learning: representations of a stochastic environment. *Theor. Popul. Biol.* **48**, 251-276.

Brachman, R. J. (2002). Systems that know what they're doing. *IEEE Intelligent systems*. 6. 67-71.

Burtsev, M.S.(2004). Tracking the Trajectories of Evolution. *Artificial Life*. V. **10.** 4. 397-411.

Bush, R.R. & Mosteller, F. (1955). *Stochastic models for Learning.* New York: Wiley.

Cohen A., Ivry R. I., & Keele S. W. (1990). Attention and structure in sequence learning. *Journal of Experimental Psychology: Learning, Memory, and Cognition.* 16.

Curran T., & Keele S.W. (1993). Attentional and nonattentional forms of sequence learning. *Journal of Experimental Psychology: Learning, Memory and Cognition.* 19.

Delgado, J., Sole, R. (2000). Self-synchronization and Task Fulfilment in Ant Colonies. *Journal of Theoretical Biology,* **205**. 3. 433-441.

Dukas, R. (1998). Constraints on information processing and their effects on behavior. In: *Cognitive ecology: The Evolutionary Ecology of Information Processing and Decision Making* (Ducas, R., ed.), 89-127. Chicago: University of Chicago Press.

Frank, S. (1997). The Design of Adaptive Systems: Optimal Parameters for Variation and Selection in Learning and Development. *Journal of Theoretical Biology,* **184**. 1. 31-39.

Garey, M. & Johnson, D. (1979). Computers and Intractability: A Guide to the Theory of NP-Completeness. San Francisco: Freeman.

Handbook of Intelligent Control. Neural, Fuzzy, and Adaptive Approaches (1992). Eds. D.A. White, D.A. Sofge. N.Y.: Van Nostrand Reinhold.

Hauert, C., Stenull, O. (2002). Simple Adaptive Strategy Wins the Prisoner's Dilemma. *Journal of Theoretical Biology,* **218**. 3. 261-272.

Intelligent Control Systems: Theory and Applications (1996). Eds M.M. Gupta, N.K. Sinha. – N.Y.: IEEE Press.

Iizuka, H., Ikegami, T. (2004). Adaptability and diversity in simulated turn-taking behavior. *Artificial Life.* **10**. 361-378.

Jonker K.C., Snoep J.L., Treur J., Westerhoff H.V., Wijngraads W.C.A. (2002). Putting intentions into cell biochemistry: an artificial intelligence perspective. *Journal of Theoretical Biology*. **214.** 105-134.

Kerr, B., Feldman, M.W. (2003). Carving the cognitive niche: optimal learning strategies in homogenous and heterogeneous environments. *J. Theor. Biol.* **220**, 169-188. doi:10.1006/jtbi.2003.3146.

Luger, G.F. (1994). Cognitive Science: The Science of Intelligent Systems. San Diego and New York: Academic Press.

Luger, G.F. (2003). Artificial intelligence. Structures and strategies for complex problem solving. Fourth edition. Addison Wesley.

Maturana, H.R. and Varela, F.J. (1980). *Autopoiesis and Cognition. The realization of the living.* Dortrecht, The Netherlands: D. Reidel.

McFarland, D. (1985). *Animal Behaviour, Psychology, Ethology and Evolution.* Univ. Of Oxford, Pitman.

Melkikh, A.V. (2002). Can Organism Pick up New Valuable Information from the Environment? *Biophysics (Biofizika).* **47**. 6. 1053-1058.

Melkikh, A.V. (2005). Congenital programs of the behavior as the unique basis of the brain activity. *NeuroQuantology.* 2. 134-148.

Minsky, M.L. (1987). *The Society of Mind.* William Heinemann Ltd., London.

Papert S. (1980). Mindstorms. New York: Basic Books.

Rich E., Knight K. (1991). Artificial Intelligence. N.Y.: McGraw-Hill.

Ruiz-Mirazo, K., Moreno, A. (2004). Basic Autonomy as a fundamental step in the synthesis of life. *Artificial Life.* **10**. 235-259.

Sloman, A. (1993). *The mind as a control system.* In Hookway C. and Peterson D., editors, Phylosophy and the Cognitive Sciences, 69-110. Cambridge University Press.

Staddon, J.E.R. (1983). *Adaptive behavior and learning.* Cambridge: Cambridge University Press.

Spier, E. & McFarland, D. (1997). Possibly Optimal Decision-Making under Self-sufficiency and Autonomy. *Journal of Theoretical Biology,* **189**. 3. 317-331.

Thompson, P., Cannon, T.D., Toga, A.W. (2002). Mapping genetic influences on human brain structure. *Annals of Medicine, 34*, 523-536.

Winston P.H. (1992). Artificial Intelligence. Addison-Wesley, Reading, Massachusetts, Third Edition.